\documentclass[conference]{IEEEtran}
\IEEEoverridecommandlockouts

\usepackage{caption}
\usepackage{graphicx}
\usepackage{amsmath,amssymb,thmtools,enumerate}
\usepackage{setspace}
\usepackage{booktabs}
\usepackage[caption=false,font=footnotesize]{subfig}
\usepackage{balance}

\usepackage[usenames,dvipsnames,svgnames,table]{xcolor}
\usepackage[colorlinks=true,pdfpagemode=UseNone,citecolor=black,linkcolor=black,urlcolor=BrickRed,
pagebackref]{hyperref}

\usepackage{etoolbox}
\makeatletter
\patchcmd{\@begintheorem}{\textit}{\textbf}{}{}
\makeatother

\newtheorem{problem}{Problem}

\newtheorem{remark}{Remark}

\newcommand{\Dcal}{\mathcal{D}}

\newcommand{\Xcal}{\mathcal{X}}

\newcommand{\m}{\mathop{\mathrm{m}}}
\newcommand{\dBm}{\mathop{\mathrm{dBm}}}
\newcommand{\Hz}{\mathop{\mathrm{Hz}}}
\newcommand{\GHz}{\mathop{\mathrm{GHz}}}
\newcommand{\MHz}{\mathop{\mathrm{MHz}}}

\newcommand{\SEth}{\mathop{\mathrm{SE}(3)}}
\newcommand{\SOth}{\mathop{\mathrm{SO}(3)}}
\newcommand{\Rth}{\mathop{\mathbb{R}^3}}
\newcommand{\Rgeq}{\mathop{\mathbb{R}_{\geq 0}}}
\newcommand{\transpose}{\mathsf{T}}

\newcommand{\squeezeup}{\vspace{-4mm}}

\title{\LARGE \bf
A Radio-Inertial Localization and Tracking System with BLE Beacons Prior Maps
}

\author{Maani Ghaffari Jadidi, Mitesh Patel, Jaime Valls Miro, Gamini Dissanayake\\ Jacob Biehl, and Andreas Girgensohn%
\thanks{Maani Ghaffari Jadidi is with Department of Naval Architecture and Marine Engineering, University of Michigan, Ann Arbor, MI 48109 USA {\tt\small maanigj@umich.edu}; this work was performed during the tenure of his position at the University of Technology Sydney.}%
\thanks{Mitesh Patel, Jacob Biel, and Andreas Girgensohn are with FX Palo Alto Laboratory Inc., Palo Alto, CA - 94304, USA {\tt\small \{mitesh,biehl,andreasg\}@fxpal.com}}%
\thanks{Jaime Valls Miro, and Gamini Dissanayake are with Centre for Autonomous System, Faculty of Engineering and IT, University of Technology Sydney, Ultimo, NSW 2007, Australia {\tt\small \{jaime.vallsmiro, gamini.dissanayake\}@uts.edu.au}}%
}

\begin{document}
\maketitle

\begin{abstract}
In this paper, we develop a system for the low-cost indoor localization and tracking problem using radio signal strength indicator, Inertial Measurement Unit (IMU), and magnetometer sensors. We develop a novel and simplified probabilistic IMU motion model as the proposal distribution of the sequential Monte-Carlo technique to track the robot trajectory. Our algorithm can globally localize and track a robot with \emph{a priori} unknown location, given an informative prior map of the Bluetooth Low Energy (BLE) beacons. Also, we formulate the problem as an optimization problem that serves as the Back-end of the algorithm mentioned above (Front-end). Thus, by simultaneously solving for the robot trajectory and the map of BLE beacons, we recover a continuous and smooth trajectory of the robot, corrected locations of the BLE beacons, and the time-varying IMU bias. The evaluations achieved using hardware show that through the proposed closed-loop system the localization performance can be improved; furthermore, the system becomes robust to the error in the map of beacons by feeding back the optimized map to the Front-end.
\end{abstract}

\begin{IEEEkeywords}
Localization, SLAM, Particle Filtering, Nonlinear Filtering, Probability and Statistical Methods, Range Sensing, Radio-Inertial Localization and Tracking.
\end{IEEEkeywords}

\IEEEpeerreviewmaketitle

\section{INTRODUCTION}

Indoor positioning systems are crucial for applications such as asset tracking and inventory management. Such systems can also increase the performance of first responders. The Visual-Inertial Navigation Systems (VINS) provide reliable solutions in both indoors and outdoors~\cite{lupton2012visual,forster2016manifold}. However, VINS often rely on proper lighting and rich visual information streams. Furthermore, using cameras can raise privacy concerns~\cite{rueben2016evaluation} which can limit the application of such systems. As an alternative or (as we prefer) a complementary solution, indoor positioning systems based on standard wireless communication technologies have also been studied extensively~\cite{liu2007,yanying2009}. In indoors, radio signals are severely impacted due to shadowing and multipathing effects~\cite{rappaport1996wireless,goldsmith2005wireless} which make the available wireless-based positioning systems less accurate ($1-10\m$)~\cite{liu2007,evall2017}.

Wireless Local Area Network (WLAN) and Bluetooth Low Energy (BLE) technologies are widespread and ubiquitous. Thus, we focus our attention on Radio Signal Strength Indicator (RSSI) available through WLAN and BLE broadcasts. In our previous work~\cite{jadidi2017gaussian}, we developed an RSSI-based indoor localization framework embedded with an online observation classifier to localize a smartphone user or a robot in a given environment. Similar to~\cite{jadidi2017gaussian}, we use Sample Importance Resampling (SIR) filter (or particle filter) embedded with the systematic resampling algorithm as it is suitable to deal with global uncertainty, nonlinear observation space, and multi-modal posterior densities~\cite{doucet2001sequential,ristic2004beyond,thrun2005probabilistic}.

\begin{figure}
\centering
\includegraphics[width = .9\columnwidth, trim={2.0cm 1.5cm 2.0cm 1.25cm},clip]{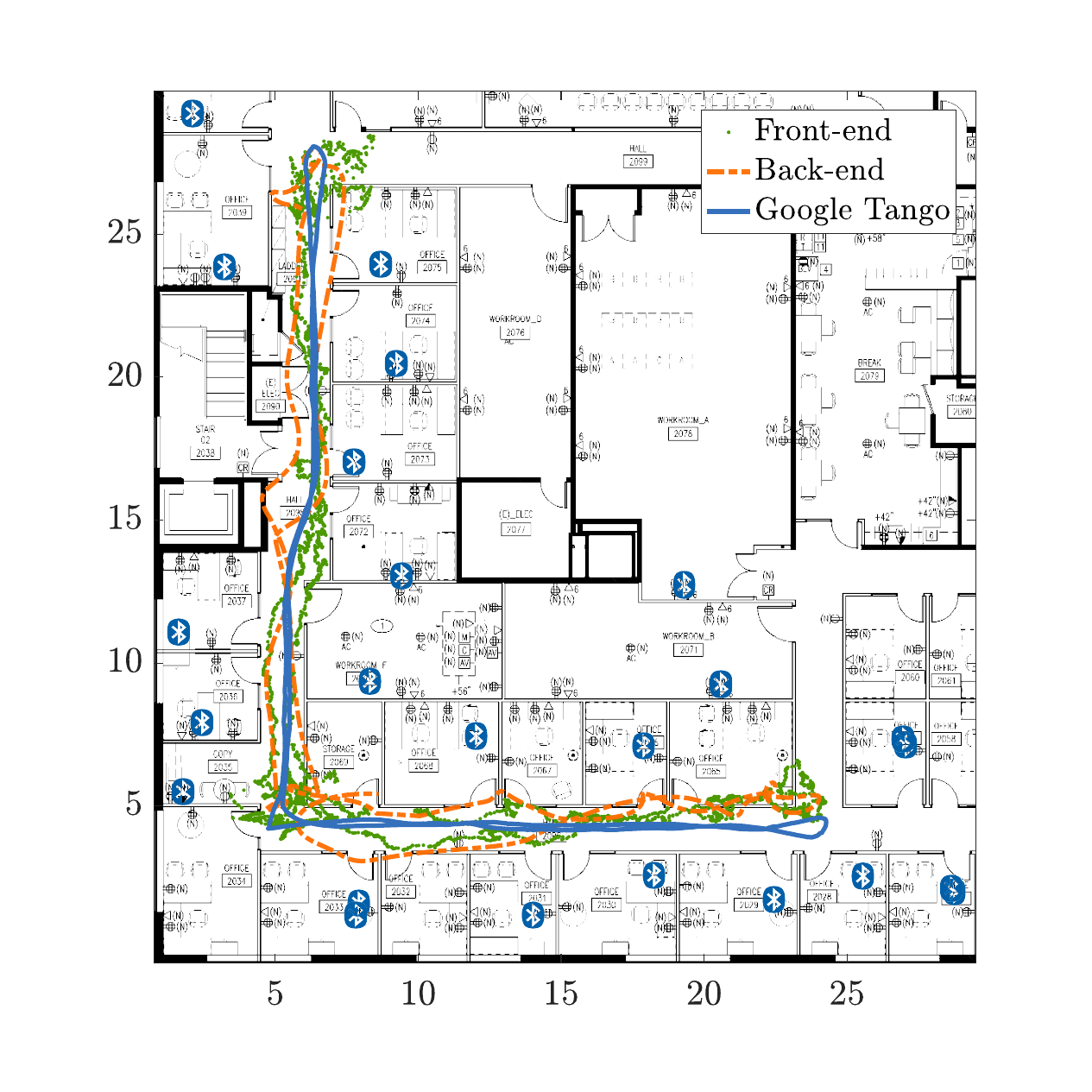}
\caption{The localization and tracking results in an office environment populated with BLE beacons. The Front-end outputs are the PF estimates and discrete. The Back-end trajectory is continuous and smooth due to the motion constraints enforced by IMU measurements. Google Tango~\cite{googletango} is used as a proxy for the ground truth trajectory. The total traveled distance is more than $100 \m$. The map of the BLE beacons is known \emph{a priori}. The Map dimensions are in meters.}
\label{fig:fx_office}
\squeezeup
\end{figure}

In this paper, we propose a radio-inertial localization and tracking system that exploits BLE, Inertial Measurement Unit (IMU), and magnetometer sensors with the quality available in standard smartphones; an illustrative example of the achieved results is shown in Figure~\ref{fig:fx_office}. 
This paper offers the following contributions.
We propose a novel and simplified probabilistic IMU motion model that serves as the proposal distribution of the Particle Filter (PF) algorithm. The probabilistic IMU motion model enables the localization algorithm to exhibit a probabilistically sound predictive behavior and track the robot trajectory (Front-end). We then simultaneously solve for the robot trajectory, the map of BLE beacons, and the time-varying IMU bias using incremental Smoothing And Mapping (iSAM)~\cite{kaess2008isam,kaess2012isam2,dellaert2012factor} and IMU-preintegration technique~\cite{forster2016manifold} (Back-end); and develop the entire radio-inertial localization and tracking framework and show that by a closed-loop architecture, Figure~\ref{fig:system}, the overall system performance can be improved according to the cumulative distribution function of the localization error~\cite{evall2017}. In addition, the system becomes robust to the error in the map of beacons by feeding back the optimized map to the Front-end. Finally, we provide experimental evaluations along reasonably long trajectories for indoor environments.

\subsection{Outline}
A review of related works is given in the following section. Section~\ref{sec:formulation} describes the problem statement and formulation. The proposed system overview is explained in Section~\ref{sec:sysoverview}; followed by presenting the probabilistic IMU motion model in Section~\ref{sec:imumodel}. Experimental results as well as discussions on limitations of the proposed framework are presented in Section~\ref{sec:result}. Finally, Section~\ref{sec:conclusion} concludes the paper and provides ideas as future work.

\subsection{Notation}
Matrices are capitalized in bold, such as in $\mathbf{X}$, and vectors are in lower case bold type, such as in $\mathbf{x}$. Vectors are column-wise and $1\colon n$ means integers from $1$ to $n$. The Euclidean and Frobenius norms are shown by $\lVert \cdot \rVert$ and $\lVert \cdot \rVert_{\mathrm{F}}$, respectively. $\lVert \mathbf{e} \rVert_{\boldsymbol \Sigma}^2 \triangleq \mathbf{e}^{\mathrm{T}} \boldsymbol \Sigma^{-1} \mathbf{e}$. Random variables, such as $X$, and their realizations, $x$, are sometimes denoted interchangeably. $x^{[i]}$ denotes a reference to the $i$-th element of the variable. An alphabet such as $\mathcal{X}$ denotes a set. The $n$-by-$n$ identity and zero matrices are denoted by $\mathbf{I}_{n}$ and $\mathbf{O}_{n}$, respectively. $\mathbf{0}_n$ denotes a vector of zeros of size $n$. $\mathrm{vec}(x^{[1]},\dots,x^{[n]})$ denotes a vector such as $\mathbf{x}$ constructed by stacking $x^{[i]}$, $\forall i \in \{1\colon n\}$.

\section{Related Work}

WiFi or radio signal fingerprint-based indoor positioning has become the standard approach for commercial applications~\cite{liu2007,yanying2009,faragher2015location,he2016wi,cooper2016}. Such systems provide accuracies from $1-10 \m$ while they require an offline survey of the radio signal strength map. Furthermore, these systems enforce strong assumptions such as static environment and limited or slow user movements. For a recent survey see~\cite{he2016wi} and references therein. Another common technique is the angle of arrival estimation using a phased array. In~\cite{xiong:2013}, multiple WiFi access points compute angle of arrival information and aggregate them to estimate the client's location. Angle of arrival estimation is also used for localizing RFID transponders~\cite{azzouzi:2011}, but only in areas measuring a few meters. Utilizing a phased array to determine the angle of arrival is challenging for smartphone-based hardware devices as the orientation of the antennas is unknown and not accessible.

On the other end of the spectrum, Pedestrian Dead Reckoning (PDR) is a popular technique which senses user motion to perform navigation.
In PDR systems, either IMU sensors are handheld~\cite{ranko2016,zhang:2013} or mounted on different body parts, e.g.\@ foot mounted~\cite{nilsson:2014}, or torso-waist mounted~\cite{perttula:2014}. Unfortunately, these systems cannot offer a generalized solution and, for example, a person in a wheelchair cannot benefit from them.

In~\cite{zhuang:2016,corrales:2008,zou:2017}, to overcome the drawbacks encountered when using Radio Frequency (RF) or IMU sensors individually for localization, the combination of both RF and IMU sensors has been used. In~\cite{zou:2017}, PDR and WiFi fingerprinting is fused. Beacon scans occasionally correct the PDR drift. It appears that the PDR results dominate this approach with WiFi fingerprinting and Beacon scans to correct the drift. In~\cite{zhuang:2016}, a step-detection strategy is used as the motion model of an Extended Kalman Filter (EKF) while WiFi signals provide measurements. EKF is a single hypothesis filter and cannot solve the global localization or the so-called kidnapped robot problem~\cite[page 274]{thrun2005probabilistic}. Furthermore, the radio signal propagation is more likely to follow a log-normal distribution than a Gaussian~\cite{rappaport1996wireless,goldsmith2005wireless,jadidi2017gaussian}.

In this work, we bring the advances in Simultaneous Localization And Mapping (SLAM)~\cite{cadena2016past} to efficiently solve the indoor localization and tracking problem using sensors available in smart handheld devices. The main features that distinguish this work from the available radio signal-based indoor positioning literature are as follows. We develop an adaptive (online) system that does not require the tedious process of fingerprinting (site survey); hence, our approach is more scalable. We use the underlying dynamical system of the IMU sensor as the motion model, and our system is robust to outliers and occasional lack of informative observations, due to the multi-hypothesis nature of the proposed Front-end. Moreover, the system is robust to the error in the map of beacons due the SLAM back-end.

\begin{figure*}[t]
\centering
\includegraphics[width = 1.8\columnwidth, trim={0cm 0cm 0cm -1.5cm},clip]{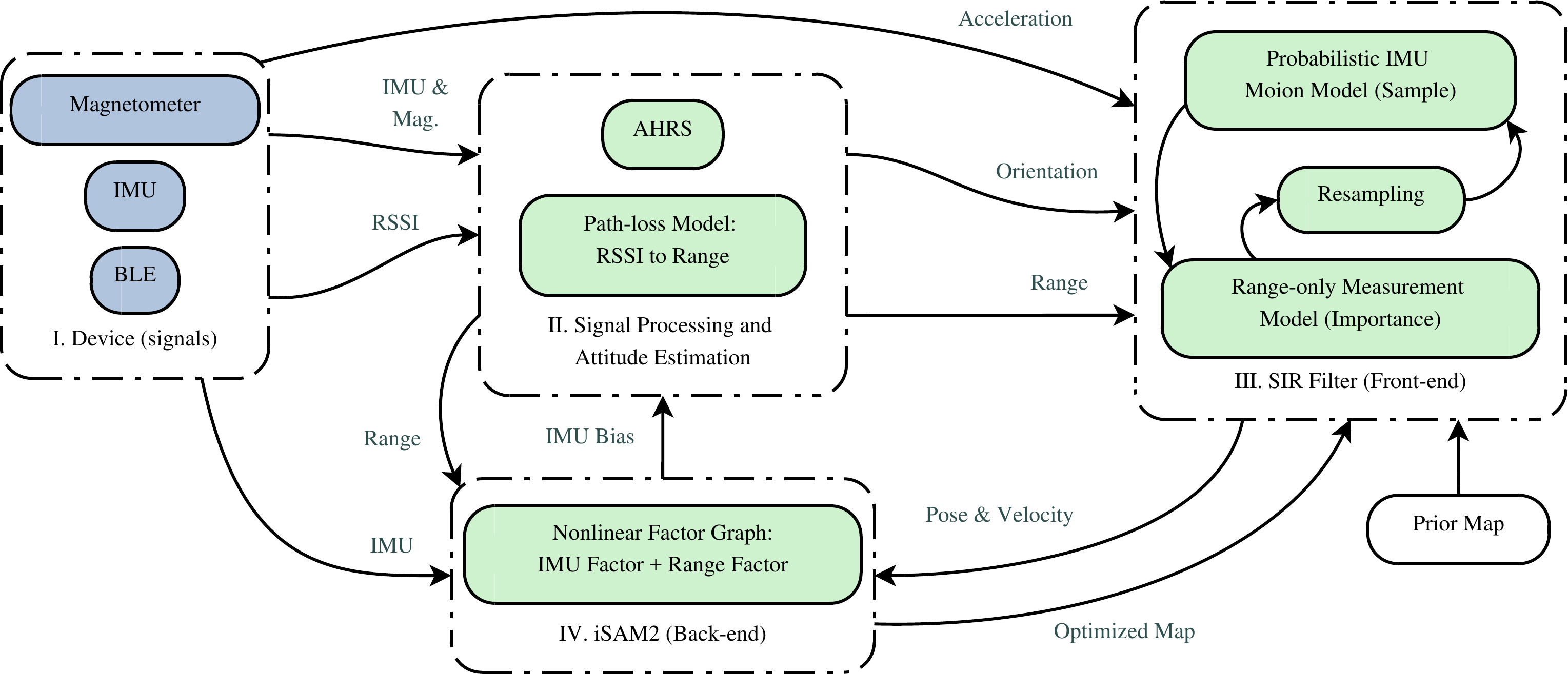}
\caption{The proposed radio-inertial localization and tracking system architecture. The Front-end position and velocity estimates are used to initialize the back-end graph nodes. Back-end provides feedback to the AHRS and Front-end in the form of time-varying IMU bias estimates and the optimized map of BLE beacons, respectively.}
\label{fig:system}
\squeezeup
\end{figure*}

\section{Problem Formulation}
\label{sec:formulation}

Let $\mathbf{x}_t \in \SEth$ be the device/robot pose at time $t$ which consists of a rotation matrix $\mathbf{R}_t \in \SOth$ and a position vector \mbox{$\mathbf{p}_t \in \Rth$}; and \mbox{$\dot{\mathbf{p}}_t \in \Rth$} denotes the corresponding velocity. The device is initially located at $\mathbf{x}_0$ which is unknown. Let \mbox{$\mathcal{L} \triangleq \{\mathbf{l}^{[j]} \in \Rth |j=1:n_l\}$} be a set of BLE beacons locations where an informative prior over any $\mathbf{l}^{[j]}$ is given. Let \mbox{$\mathcal{Z}_t \subset \Rgeq$} be the set of possible range measurements (converted RSSI) at time $t$. The probabilistic measurement model $p(z_t|\mathbf{x}_t, \mathbf{l})$ is a Gaussian conditional probability distribution that represents the likelihood of range measurements. The IMU and magnetometer measurements at time $t$ include a vector of angular velocity $\boldsymbol \omega_t \in \Rth$ and an acceleration vector $\mathbf{a}_t \in \Rth$, and a vector of local magnetic field \mbox{$\mathbf{m}_t \in \Rth$}, respectively. 

Furthermore, the control action $\mathbf{u}_t \in \mathcal{U}_t$ and the action set $\mathcal{U}_t$ have to be defined possibly based on the IMU and magnetometer measurements. Note that the purpose of control actions here is the prediction of the device motion rather than actively controlling its behavior. Therefore, here, control actions are proprioceptive measurements. The ultimate goal is to estimate the device initial pose (global localization) and trajectory including its position and orientation, given noisy range, IMU, and magnetometer measurements. To this end, we break the original problem into the following sub-problems.

\begin{problem}[Probabilistic motion model]
\label{prob:imumm}
Given the action set $\mathcal{U}_t$ and $\mathbf{u}_t \in \mathcal{U}_t$, find the discrete stochastic dynamics that describes transition equation $p(\mathbf{x}_t|\mathbf{x}_{t-1}, \mathbf{u}_{t-1})$.
\end{problem}

\begin{problem}[Global localization and tracking]
\label{prob:loc}
Let \mbox{$z_{1:t} \triangleq \{z_{1},...,z_{t}\}$} be a sequence of range measurements up to time $t$. Let $\mathbf{x}_t$ be a Markov process of initial distribution $p(\mathbf{x}_0)$ and transition equation $p(\mathbf{x}_t|\mathbf{x}_{t-1}, \mathbf{u}_{t-1})$. Given $p(z_t|\mathbf{x}_t, \mathbf{l})$, estimate recursively in time the posterior distribution $p(\mathbf{x}_{0:t}|z_{1:t}, \mathbf{l})$. 
\end{problem}

\begin{problem}[Radio-inertial SLAM]
\label{prob:slam}
Given all measurements up to time $t$, estimate the joint smoothing distribution $p(\mathbf{x}_{0:t}, \mathbf{l}, \boldsymbol \theta_{1:t}|z_{1:t}, \mathbf{u}_{1:t})$; where $\boldsymbol \theta_{1:t}$ are the, possibly time-varying, system calibration parameters.
\end{problem}

We solve Problem~\ref{prob:imumm} by developing the probabilistic IMU motion model and embed it into the SIR filter as the proposal distribution to solve Problem~\ref{prob:loc}. In Problem~\ref{prob:slam}, we solve the SLAM problem to estimate the smoothing distribution of the device trajectory, the optimized map, as well as time-varying IMU bias. We note that solving the SLAM problem has the following advantages:
\begin{enumerate}
\item We relax the need for an exact prior map of beacons and, given sufficient range measurements, any misalignment in the beacon placement can be recovered.
\item Through joint parameter and state estimation, we can recover system calibration parameters in a computationally affordable manner.
\end{enumerate}

\begin{remark}
The magnetometer sensor is almost available on all smart handheld devices alongside the IMU. However, it is possible to remove magnetometer measurements from the problem formulation, while the problem remains solvable. In this case, the attitude estimation accuracy might degrade. 
\end{remark}

\section{System Overview and Architecture}
\label{sec:sysoverview}

The proposed localization system is suitable for GPS-denied indoor environments and consists of two main layers which we call them Front-end and Back-end. The entire system architecture is shown in Figure~\ref{fig:system}. The necessary modules can be split up into four parts; in the following, we describe each module.

\subsection{Sensors and Raw Signals}
\label{subsec:sensors}
Standard and commercially available smartphones typically are equipped with Bluetooth transceiver, IMU, and magnetometer sensors. This module receives raw signals. The BLE operates in the $2.4\GHz$ license-free band and uses 40 channels each with a width of $2\MHz$; the ideal sampling rate of RSSI is about $10\Hz$~\cite{faragher2015location}. The IMU sensor includes a 3-axis accelerometer, which measures the linear acceleration, and a 3-axis gyroscope, that measures the angular velocity, and typically has a sampling rate higher than $100\Hz$. The magnetometer measures the strength and direction of the local magnetic field and has a sampling rate of about $50\Hz$.

\subsection{Signal Processing and Attitude Estimation}
\label{subsec:ahrs}

In this module, we apply median filter to IMU and magnetometer signals to remove noise. However, the BLE signals depend on the availability of a link between the beacon and the receiver; as such, the filtering is only possible when there is a sufficient sequence of the RSSI. After this pre-processing step, we use the simplified free space path-loss model to convert RSSI to range measurements. For details of this step see~\cite{jadidi2017gaussian}. In this work, we clamp range measurements by discarding any range value greater than $10\m$. This is the known effective range of the BLE technology~\cite{bleprotocol2010,jadidi2017gaussian}.

Using IMU and magnetometer measurements, we solve the Attitude and Heading Reference System (AHRS)~\cite{euston2008complementary,madgwick2011estimation} problem to estimate the device orientation. Ideally, the magnetometer senses the direction of the Earth gravitational field which provides the North-East-Down coordinate system. However, if the local gravitational field is affected by external disturbances such as the structure of the building, this reference system might not be valid anymore and needs to be verified. Upon availability of the Back-end module, as shown in Figure~\ref{fig:system}, AHRS can compensate IMU bias which improves the orientation estimation.

\subsection{Front-end}
\label{subsec:frontend}
In the Front-end we use the sequential Monte-Carlo or PF technique known as SIR filter to solve the problem of global localization and tracking the device trajectory. Given the current device orientation, we sample from the IMU dynamics to generate particles along the moving direction. Given a prior map of the BLE beacons and using range measurements converted from BLE RSSI, we compute importance weights of particles. In this work, we use a range-only measurement model with additive white Gaussian noise. Finally, in the Resampling step particles with higher weights are replicated and all weights are set uniformly. The Sample, Importance, and Resampling steps are repeated sequentially.

Since the IMU has a higher frequency than the BLE receiver, the AHRS solver runs with a higher frequency to use all available measurements which in turn improves the orientation estimation accuracy. We sample from the IMU motion model using the latest device orientation with a frequency that is adaptable to the available computational resources (usually the BLE RSSI sampling rate). In this way, without discarding any sensory information, we can track the device within the Particle Filter framework.

\subsection{Back-end}
\label{subsec:backend}
The Back-end includes an incremental optimization (smoothing) algorithm to minimize the error in Front-end estimates as well as estimation of the system calibration parameters. Let $\Xcal_t \triangleq (\mathbf{x}_t, \dot{\mathbf{p}}_t, \mathbf{l}, \boldsymbol \theta_t)$ be the state variables tuple at any time step $t$. Let $\Dcal_t$ be the set of all range measurements at time step $t$ and all IMU measurements from time step $t-1$ to $t$. The joint probability distribution of the SLAM problem, Problem~\ref{prob:slam}, by assuming the measurements are conditionally independent and are corrupted by additive zero mean white Gaussian noise can be written as follows~\cite{dellaert2006square}:

\begin{equation}
\label{eq:slam1}
  p(\Xcal_{0:T},\Dcal_{1:T}) = p(\Xcal_0) \prod_{t=1}^T p(\Dcal_t|\Xcal_t)
\end{equation}
\begin{equation}
\label{eq:slam2}
  \log p(\Xcal_{0:T},\Dcal_{1:T}) = \log p(\Xcal_0) + \sum_{t=1}^T \log p(\Dcal_t|\Xcal_t)
\end{equation}

Given that $p(\Xcal_{0:T}|\Dcal_{1:T}) \propto p(\Xcal_0) \prod_{t=1}^T p(\Dcal_t|\Xcal_t)$, the \emph{maximum-a-posteriori} estimate of $\Xcal_{0:T}$ can be computed by solving the following nonlinear least-squares problem:
\begin{align}
\label{eq:slam3}
  \nonumber \Xcal_{0:T}^\star = & \underset{\Xcal_{0:T}}{\operatorname{arg\ min}} \ -\log p(\Xcal_{0:T}|\Dcal_{1:T}) \\
    & = \underset{\Xcal_{0:T}}{\operatorname{arg\ min}} \lVert \mathbf{r}_0 \rVert_{\boldsymbol \Sigma_0}^2 + \lVert \mathbf{r}_{0:T} \rVert_{\boldsymbol \Sigma_\mathbf{r}}^2
\end{align}
where $\mathbf{r}_t$ denotes the residual term which is the error between measurements and their corresponding nonlinear models, i.e.\@ range-only measurement model and preintegrated IMU measurement model as described in~\cite{forster2016manifold}; $\boldsymbol \Sigma_0$ and $\boldsymbol \Sigma_r$ denote the corresponding measurement noise covariances.

Therefore, the optimization algorithm simultaneously solves for time-varying IMU bias, the feature map of BLE beacons, and the entire device trajectory. The resultant trajectory is smooth and continuous, unlike PF output, due to the motion constraints enforced by the IMU measurements. Furthermore, the estimated IMU bias is fed to the AHRS algorithm to improve the attitude estimation; and the optimized BLE map is used in the measurement update step of the SIR filter in the Front-end. This closed-loop architecture improves the Front-end performance as we will see in Section~\ref{sec:result}. In this work, we use iSAM2 algorithm~\cite{kaess2012isam2} and the GTSAM library~\cite{dellaert2012factor} as the Back-end solver.

\section{Probabilistic IMU Motion Model}
\label{sec:imumodel}
In this section, we describe the proposed simplified method to incorporate the IMU dynamics into the PF that does not require sampling from the full 6-dimensional state (pose). In general, sampling methods tend to become inefficient and computationally intensive as the dimension of the state grows. The key idea is to sample the tangential acceleration magnitude along the direction of movement and evolve the state estimate using a discrete-time stochastic dynamical system that corresponds to the IMU dynamics. 

At any time $t$, using the AHRS solver the current orientation estimate, $\hat{\mathbf{R}}_t$, is given. Therefore, by knowing the orientation, the system dynamics can be expressed using the following linear system:
\begin{equation}
\label{eq:systemdyn}
\bar{\mathbf{x}}_{t+1} = \mathbf{F}_t \bar{\mathbf{x}}_t + \mathbf{G}_t \mathbf{u}_t
\end{equation}
where \mbox{$\bar{\mathbf{x}}_t \triangleq \mathrm{vec}(\mathbf{p}_t, \dot{\mathbf{p}}_t) \in \mathbb{R}^6$} is the state vector (of Front-end), and \mbox{$\mathbf{F}_t \in \mathbb{R}^{6\times 6}$} and \mbox{$\mathbf{G}_t \in \mathbb{R}^{6\times 6}$} are system and input matrices, respectively, and using a sampling time, $t_s$, can be derived as follows.
\begin{equation}
\mathbf{F}_t = 
\begin{bmatrix}
	\mathbf{I}_3 & t_s \mathbf{I}_3 \\
	\mathbf{O}_3 & \mathbf{I}_3
\end{bmatrix}
,\quad
\mathbf{G}_t = 
\begin{bmatrix}
	0.5t_s^2 \mathbf{I}_3	& t_s \mathbf{I}_3 \\
	t_s \mathbf{I}_3	& \mathbf{I}_3
\end{bmatrix}
\end{equation}

We can compute the current tangential acceleration direction, $\hat{\mathbf{a}}_{t}$, as follows.
\begin{equation}
\label{eq:gacc}
\tilde{\mathbf{a}}_t = \hat{\mathbf{R}}_t \mathbf{a}_t + \mathbf{g}
\end{equation}
\begin{equation}
\label{eq:gaccdir}
\hat{\mathbf{a}}_{t} = \frac{\tilde{\mathbf{a}}_t}{\lVert \tilde{\mathbf{a}}_t \rVert}  \quad \lVert \tilde{\mathbf{a}}_t \rVert \neq 0
\end{equation}
where $\tilde{\mathbf{a}}_t$ is the acceleration vector in the global coordinates after compensating the gravity $\mathbf{g}$. Let \mbox{$\zeta_t \sim \mathcal{N}(\boldsymbol 0, \sigma_a)$} be the sampled tangential acceleration magnitude where $\sigma_a$ is a sufficiently large linear acceleration standard deviation that covers a typical range of activities from slow walking to sudden movements. Furthermore, to maintain the diversity of the particles, let $\boldsymbol \nu_t \sim \mathcal{N}(\boldsymbol 0, \sigma_v \mathbf{I}_3)$ be the sampled velocity vector with the isotropic covariance matrix, $\sigma_v \mathbf{I}_3$, that perturbs the position of the particles. We can then construct the input vector, \mbox{$\mathbf{u}_t \in \mathbb{R}^6$}, as \mbox{$\mathbf{u}_t = \mathrm{vec}(\zeta_t \hat{\mathbf{a}}_{t}, \boldsymbol \nu_t)$}.

Therefore, we addressed Problem~\ref{prob:imumm} using the proposed stochastic dynamical system that can serve as the transition equation $p(\bar{\mathbf{x}}_t|\bar{\mathbf{x}}_{t-1}, \mathbf{u}_{t-1})$. The proposed transition equation enables PF algorithm to track the device trajectory by predicting its motion. To address Problem~\ref{prob:loc} and \ref{prob:slam}, we use the described Front-end and Back-end, respectively. We then integrate all modules into a unified closed-loop system (Figure~\ref{fig:system}).

\begin{remark}
 We note that upon availability of the Back-end for online system calibration task, the proposed IMU motion model, and therefore the Front-end can benefit from it. Consequently, the acceleration bias, $\boldsymbol \theta_{a,t}$, can be included in~\eqref{eq:gacc}, i.e., $\tilde{\mathbf{a}}_t = \hat{\mathbf{R}}_t (\mathbf{a}_t - \boldsymbol \theta_{a,t}) + \mathbf{g}$. However, as we will show later in the presented evaluations, the model at its basic form does not depend on this parameter to provide comparable results as it is used for the sampling purpose.
\end{remark}

 One might consider sampling $\zeta_t$ from a normal distribution centered at the current measured acceleration magnitude. However, we empirically observed that this method degrades the performance of the system as the measured magnitude, without calibration, is not reliable. Another reason to sample from a zero mean distribution is the fact that the variance can be sufficiently large to cover all possible values. Theoretically speaking in particle filtering, a sampling strategy is well-behaved as long as the proposal distribution's support includes that of the true posterior distribution. Therefore, the claim for better tracking using the proposed simplified IMU model is connected to the fact that it provides a tighter support while still covers the support of the posterior distribution. Finally, we note that the proposed probabilistic IMU motion model follows the standard IMU dynamics as discussed in~\cite{forster2016manifold}. However, by providing the orientation, we propose a sampling strategy that is more computationally attractive for online applications.

\section{Experimental Results and Discussion}
\label{sec:result}

In this section, we evaluate the proposed system using hardware experiments. In the first experiment, the focus is on the localization problem where an accurate map of the beacons is given. We compare localization techniques using PF with the random walk motion model (labeled as RW), PF embedded with the proposed probabilistic IMU motion model (labeled as IMU), PF with the robot dead reckoning motion model (labeled as DR), and our proposed closed-loop system including the Front-end (labeled as F.-end) and the Back-end (labeled as B.-end). We note that the Front-end is the same as PF with the probabilistic IMU motion model and the improvement in the results is due to the proposed closed-loop architecture and exploiting the Back-end for online system calibration. In the second experiment, using the same experimental dataset, we run a Monte Carlo simulation to study the effect of map error on the performance of each system.

\subsection{Experimental Setup}
\label{subsec:setup}

We employ a wheeled mobile robot equipped with a Google Tango device~\cite{googletango} which provides the robot trajectory estimate using visual-inertial odometry. We use Tango's output as a proxy for ground truth. The mobile robot is also equipped with a Nexus6P smartphone\footnote{https://www.google.com/nexus/6p/} which collects BLE, IMU, and magnetometer sensors data. The data is published as Robot Operating System (ROS)~\cite{quigley2009ros} topics through a customized Android application developed in-house. The IMU and magnetometer sensors are sampled at $200 \Hz$ and $50 \Hz$, respectively. The BLE scans are sampled at $10 \Hz$; however, in practice, we experienced an average sampling rate of $7 \Hz$. The environment is populated with BLE beacons, as shown by the Bluetooth logo in Figure~\ref{fig:fx_office}. Furthermore, the prior map accuracy and the density of BLE beacons are sufficiently high to make initialization and tracking possible. The experiments are conducted in a research office environment partitioned into separate offices cabins and consists of traditional office furniture. The data is collected by maneuvering the robot over $100 \m$ in an office space of $40 \times 50 \m^2$. Note that data is collected in a natural setting on a normal working day and in presence of office staff members with no movement restrictions to staff members~\footnote{The dataset is available upon request. Please contact the authors.}.

\subsection{System Configuration and Initialization}
\label{subsec:sysinit}

To detect the degeneracy of PF, we calculate the effective sample size, $n_{eff}=(\sum_{i=1}^{n_p}w_t^{[i]})^{-1}$, and perform (systematic) resampling when $n_{eff} < n_{thr}$; where $n_p$ is the number of particles, $w_t^{[i]}$ is the $i$-th particle's weight, and $n_{thr}$ is a threshold $1 < n_{thr} < n_p$. The explanation and details of the used parameters are summarized in Table~\ref{tab:param}; and the random walk (constant velocity with random input) motion model is explained in~\cite{jadidi2017gaussian}. 
The DR motion model assumes a constant incremental movement along the AHRS estimated direction at each time step. This is because the wheel encoder data is not available in smartphones; however, this model does not maintain the diversity of the particles. Therefore, we introduce small velocity noise to perturb the position of the particles. The velocity noise standard deviation is set to the smallest value that leads to achieving comparable results.

\begin{figure*}[t!]
\centering
\subfloat{\includegraphics[width = 0.66\columnwidth]{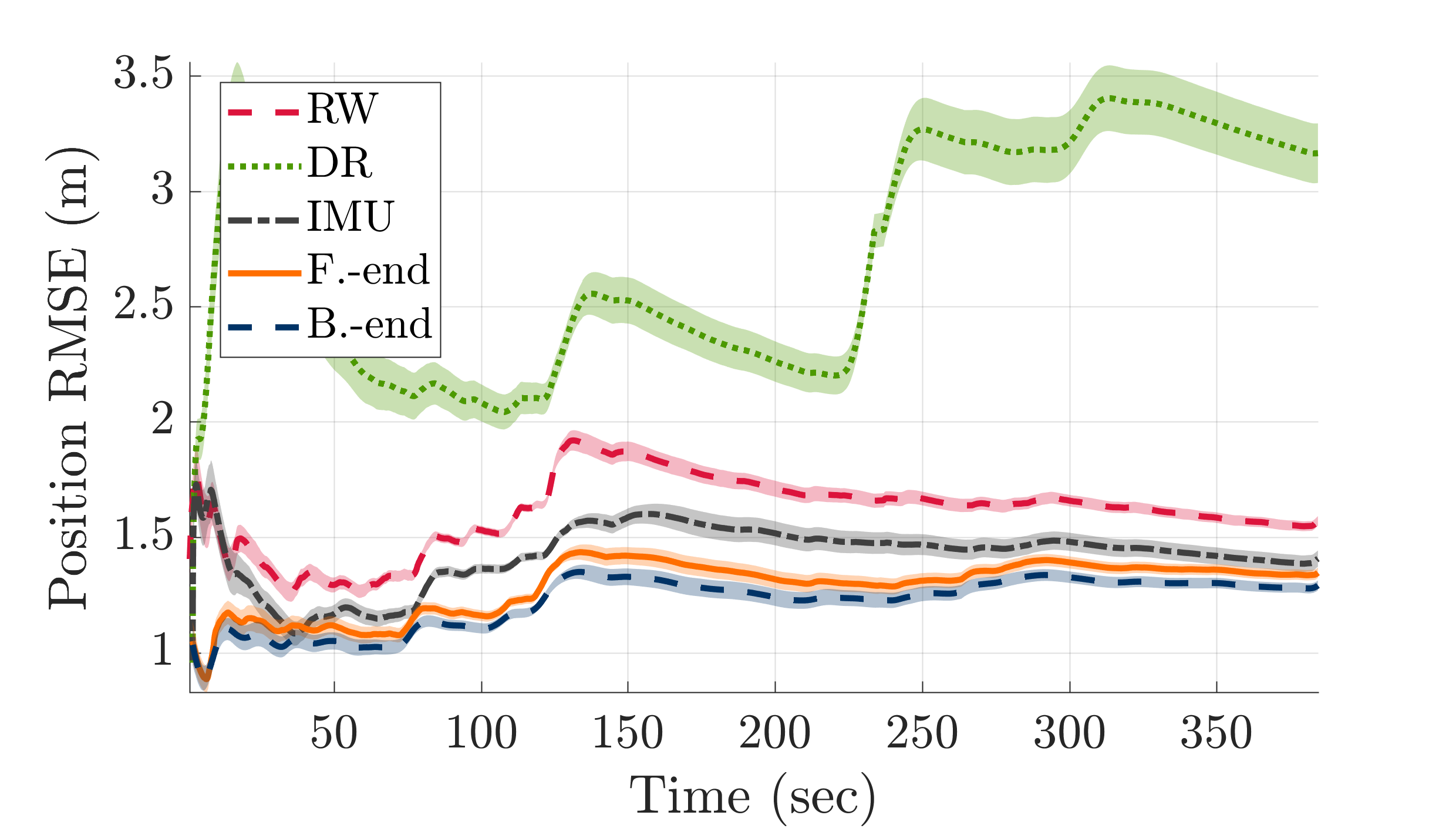}}~
\subfloat{\includegraphics[width = 0.66\columnwidth]{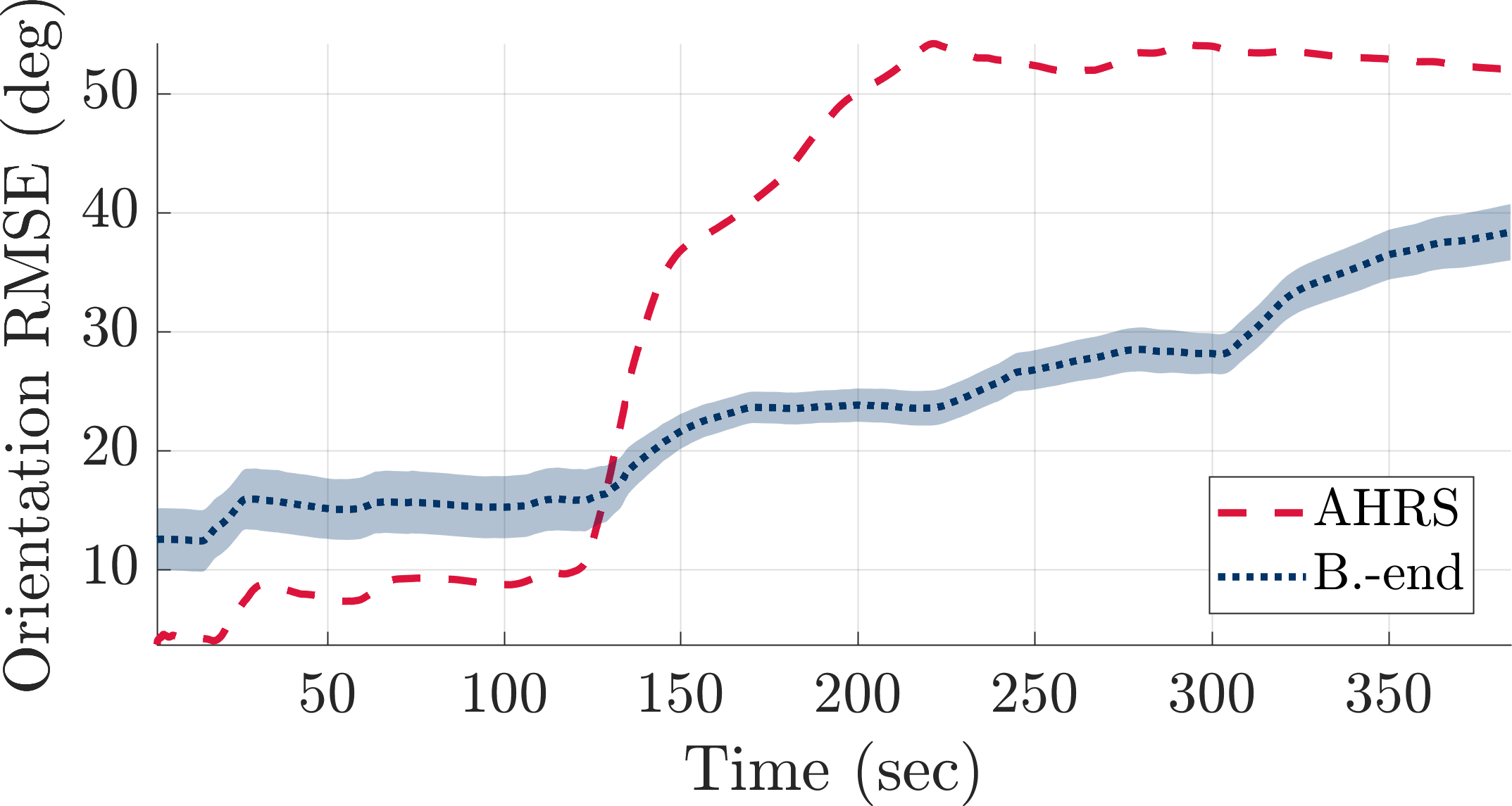}}~
\subfloat{\includegraphics[width = 0.66\columnwidth, trim={0.25cm 0.25cm 1.5cm 1cm},clip]{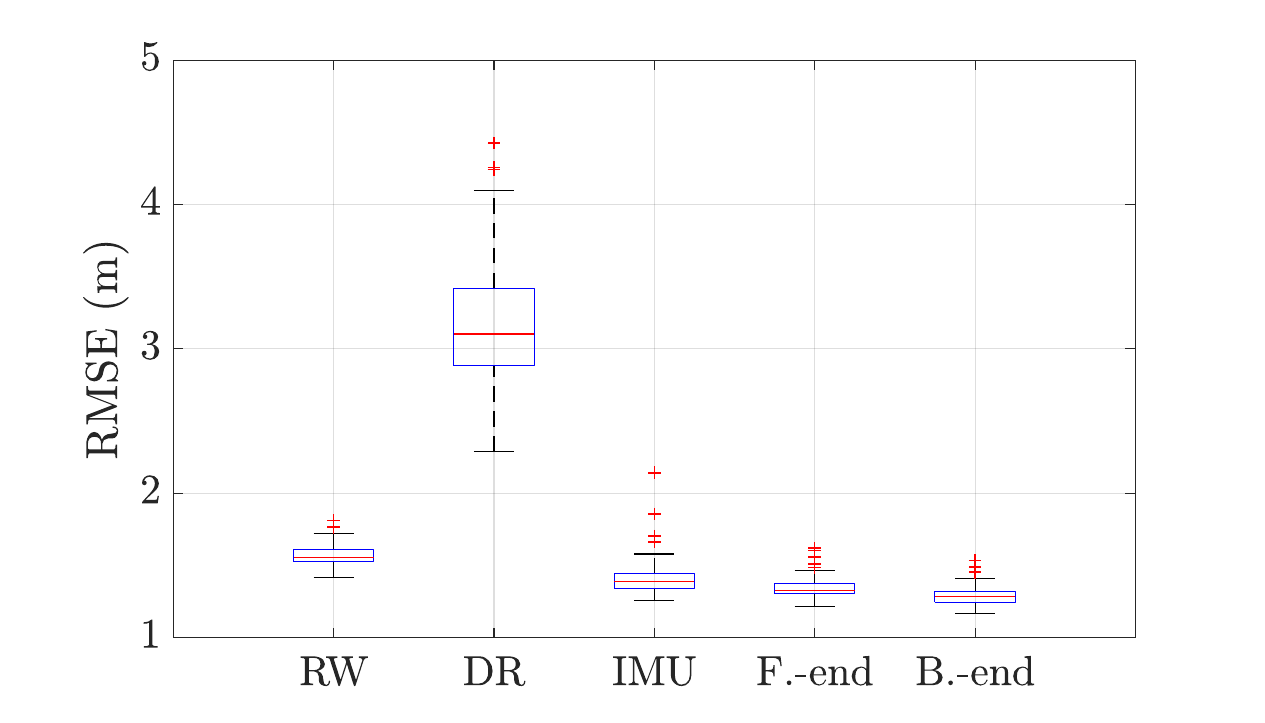}}\\ \squeezeup
\subfloat{\includegraphics[width = 0.66\columnwidth]{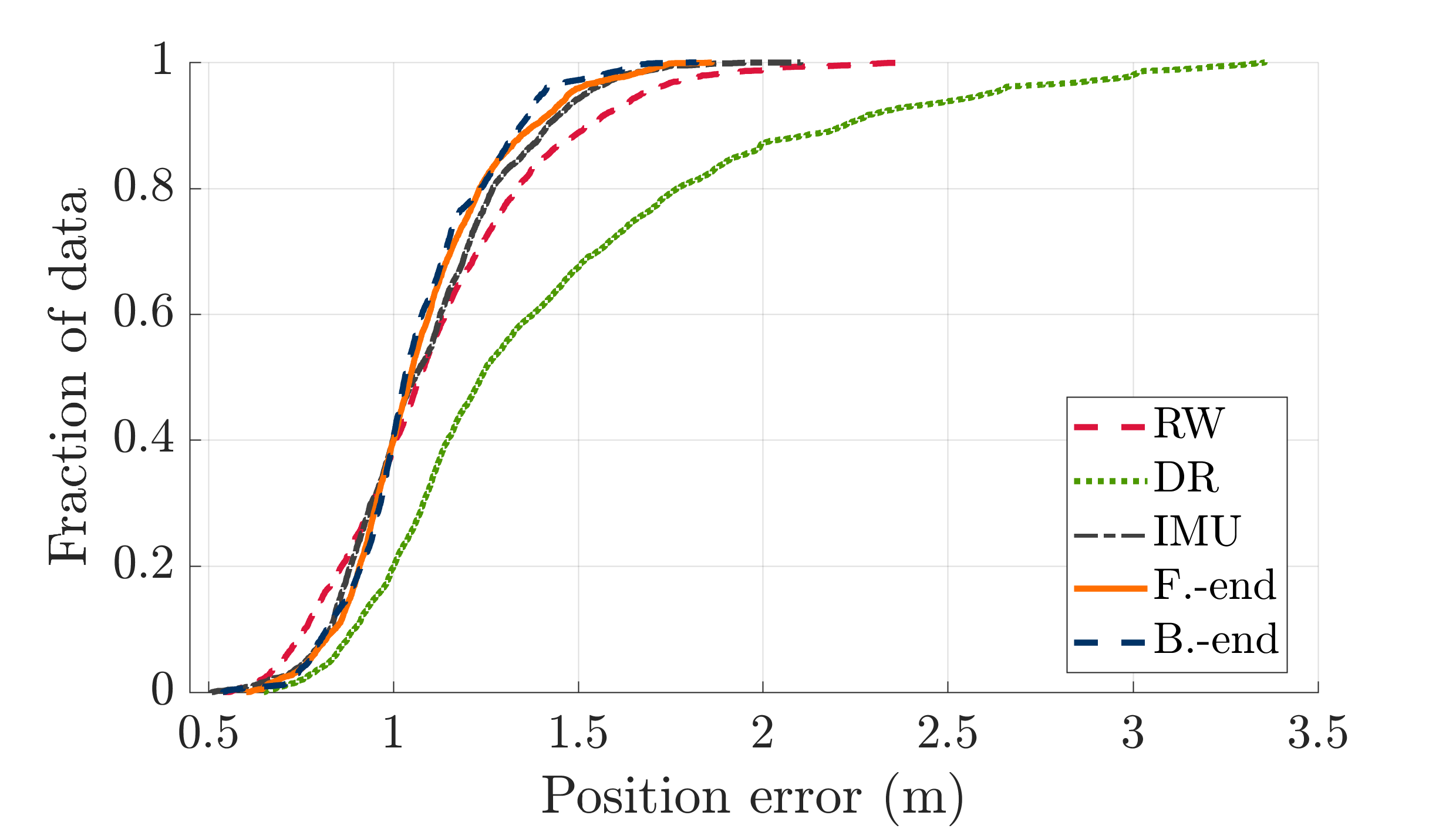}}~
\subfloat{\includegraphics[width = 0.66\columnwidth]{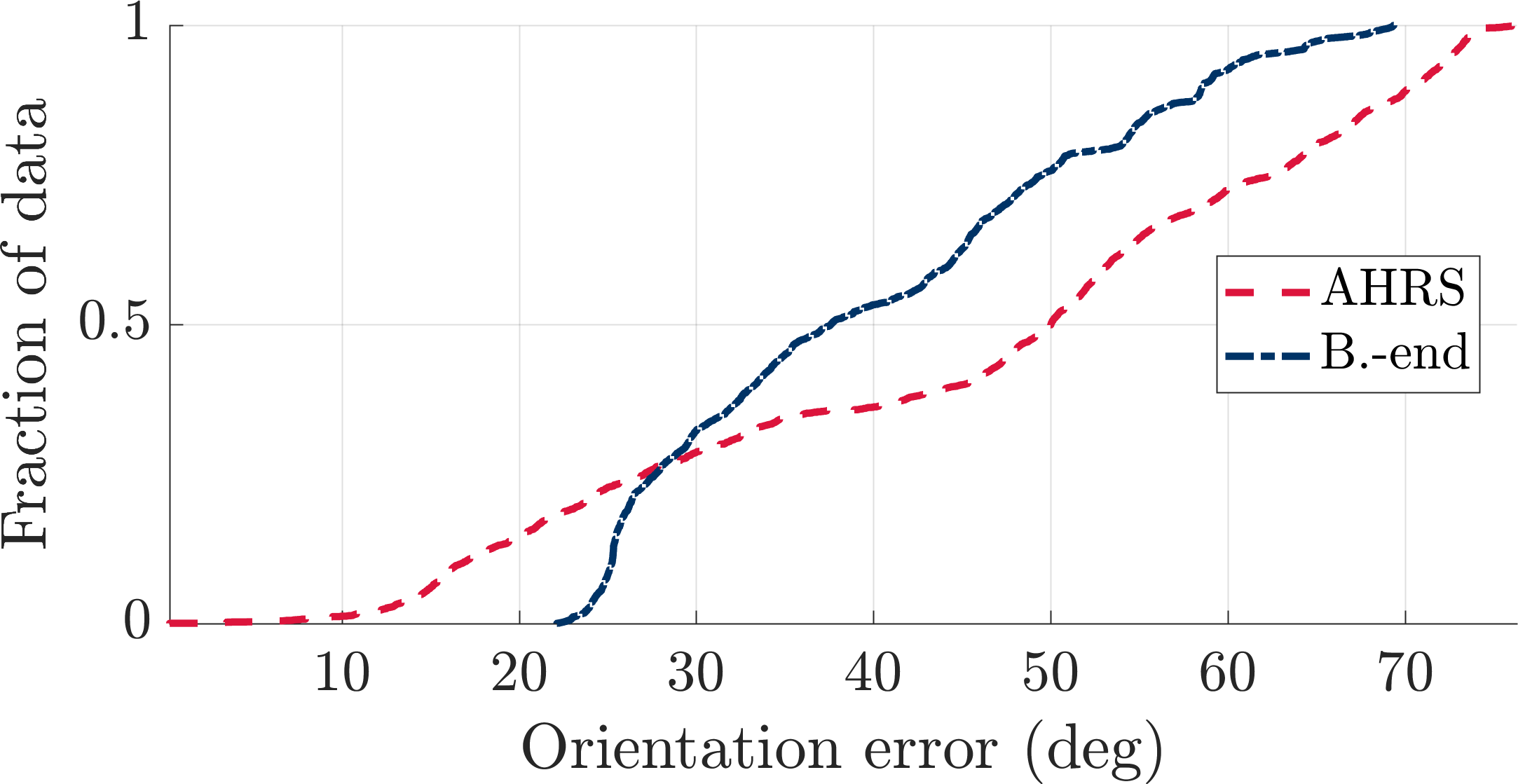}}~
\subfloat{\includegraphics[width = 0.66\columnwidth]{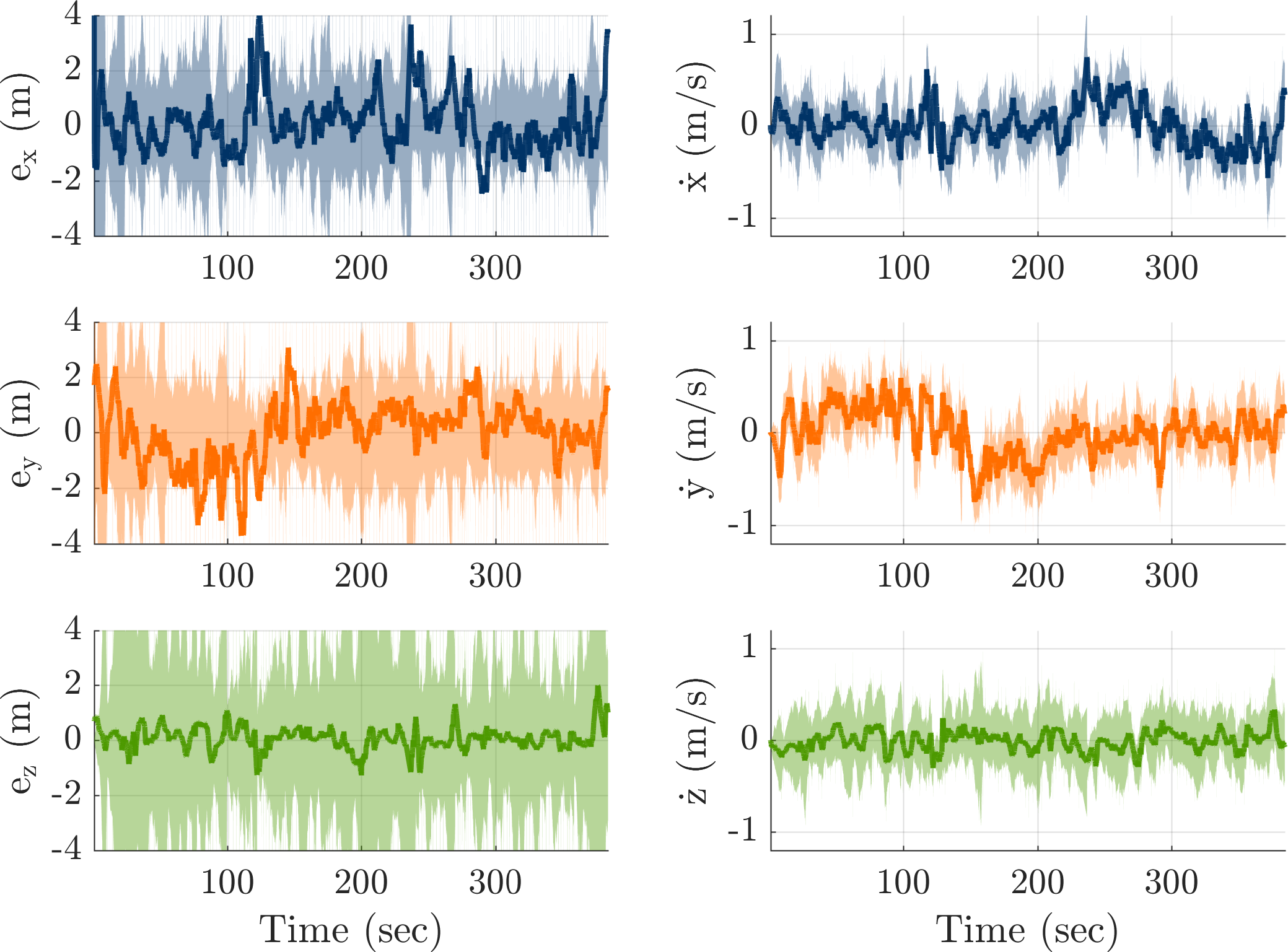}}
\caption{The experimental localization results from $100$ independent runs. From top left, respectively, figures show position RMSE (with $99\%$ confidence bounds), orientation RMSE (with $99\%$ confidence bounds), boxplot showing the statistical summary of position RMSE. From bottom left, respectively, figures show the empirical cumulative distribution functions of the compared algorithms for position error and orientation error where each curve illustrates the median of $100$ CDF. The bottom right figure shows an example of the Front-end position error and velocity estimate together with $99\%$ confidence bounds. Note that unlike the position error, the velocity uncertainty bounds are around the estimated value since we do not have access to the ground truth velocities. The computational time for processing the entire dataset in MATLAB using a laptop with an Intel Core i5 CPU is on average $42$, $60$, $66$, and $92$ seconds for RW, DR, IMU, and combined Front-end and Back-end, respectively, which are all well below the entire experiment's duration, i.e., $384$ seconds.}
\label{fig:fx_results}
\squeezeup
\end{figure*}

Our empirical observation showed that slight variations of the parameters, about $5-10\%$, do not alter the trend of the results; however, finding suitable parameters for all motion models and the path-loss model is an important part of the setup. In particular, the path-loss model parameters can be estimated as explained in~\cite{jadidi2017gaussian}. We tuned all motion models parameters manually, except the acceleration standard deviation which can be justified by considering that the average walking speed is about $1.5 \m \sec^{-1}$ and a person can reach that speed in one second. The velocity standard deviation is tuned as the secondary motion model parameter to ensure the diversity of the particles are maintained. We note that this is an important factor in all particle filtering frameworks. The first $1-2 \sec$ is sufficient for all the compared techniques to initialize, i.e., global localization. The particles are drawn uniformly within the known map area. The robot starts from the bottom left corner and moves upwards.

We clamp range measurements by discarding any range value greater than $10\m$. This is the effective range of the BLE technology~\cite{bleprotocol2010}. However, even using a smaller range does not prevent receiving non-line-of-sight RSSI measurements. In general, it is difficult to deal with non-line-of-sight observations in a computationally attractive manner, and this is the main challenge in radio signal-based indoor positioning. Therefore, we set the unusually large range standard deviation of $5 \m$ for maximum range of $10\m$ to enforce the fact that range measurements are highly noisy and inaccurate.

\begin{table}
\footnotesize
\centering
\caption{Parameters used in the experiments.}
\resizebox{\columnwidth}{!}{
\begin{tabular}{lll}
\toprule
Parameter				& Symbol	& Value \\ \midrule
\multicolumn{3}{l}{$-$ \texttt{Path-loss model parameters}~\cite{jadidi2017gaussian}:} 		\\
Attenuated transmission power  		& $a_X$		& -64.53			\\
The path-loss exponent 			& $\gamma$	& 1.72				\\  
Reference distance 			& $d_0$		& 1.78 		$\m$		\\ 
\multicolumn{3}{l}{$-$ \texttt{Range-only measurement model (Gaussian noise)}:} 		\\
standard deviation			& $\sigma_{n}$ 	& 5 		$\m$		\\
\multicolumn{3}{l}{$-$ \texttt{Random walk motion model}:} 				\\
Position standard deviation		& $\sigma_{p}$ 	& 0.1 		$\m$		\\
Velocity standard deviation		& $\sigma_{v}$ 	& 0.05 		$\m \sec^{-1}$	\\
\multicolumn{3}{l}{$-$ \texttt{Probabilistic IMU motion model}:} 			\\
Acceleration standard deviation		& $\sigma_{a}$ 	& 1.5 		$\m^2 \sec^{-1}$	\\
Velocity standard deviation		& $\sigma_{v}$ 	& 0.005 	$\m \sec^{-1}$	\\
\multicolumn{3}{l}{$-$ \texttt{Dead reckoning motion model}:} 		\\
Motion increment size			& $-$ 		& 0.075 	$\m$		\\
Velocity standard deviation		& $\sigma_{v}$ 	& 0.03 		$\m \sec^{-1}$	\\
\multicolumn{3}{l}{$-$ \texttt{Particle filter}:} 					\\
Number of particles			& $n_p$		& 300				\\ 
Resampling threshold			& $n_{thr}$	& 60				\\
\multicolumn{3}{l}{$-$ \texttt{BLE Beacon Parameters}:} 					\\
Transmission Power			& $-$		& +4 		$\dBm$ 		\\ 
Broadcasting Frequency			& $-$		& 10 		$\Hz$		\\ \bottomrule
\end{tabular}
\label{tab:param}}
\squeezeup
\end{table}

\begin{figure*}[t!]
\centering
\subfloat{\includegraphics[width = 0.66\columnwidth]{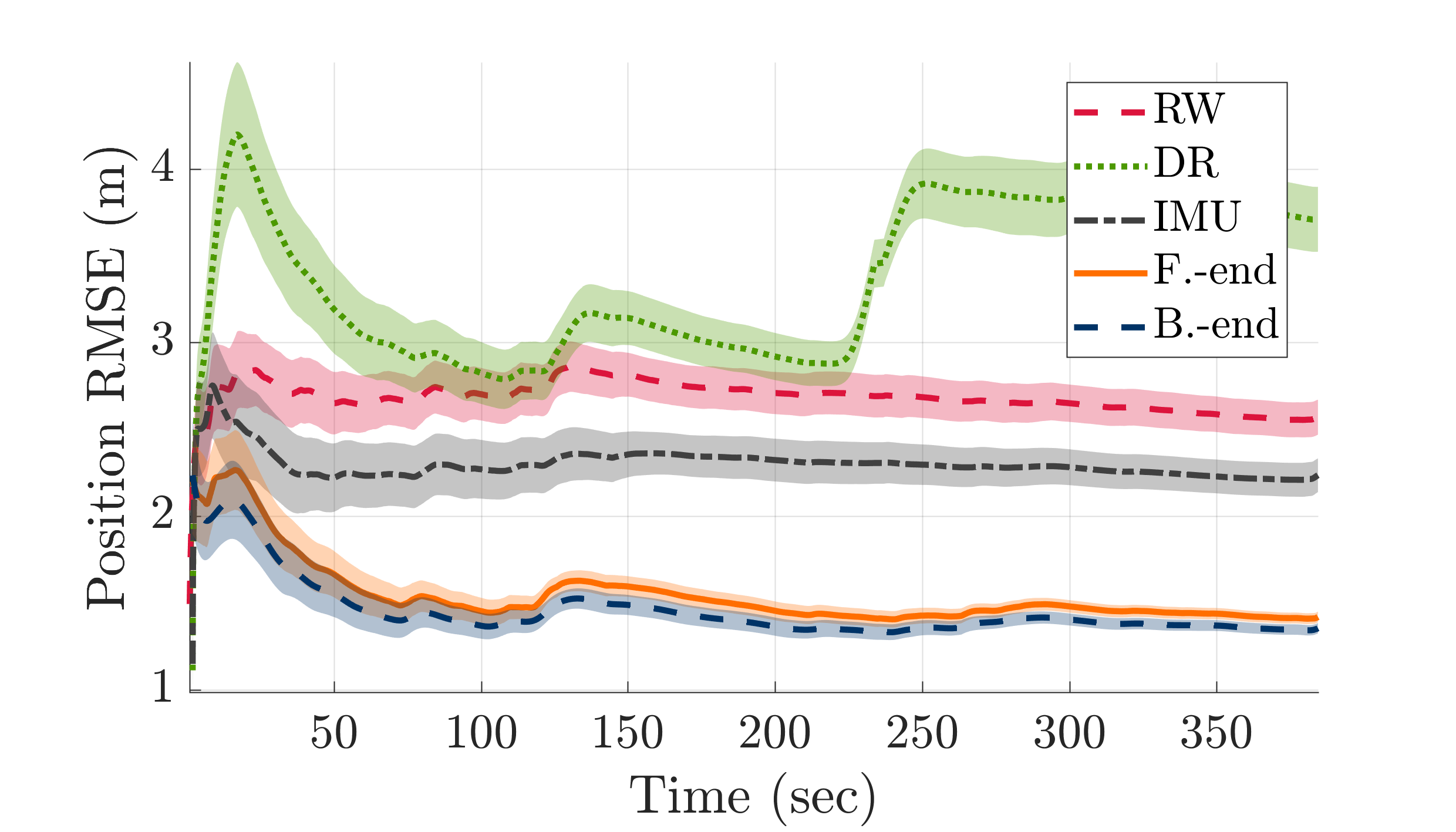}}~
\subfloat{\includegraphics[width = 0.66\columnwidth]{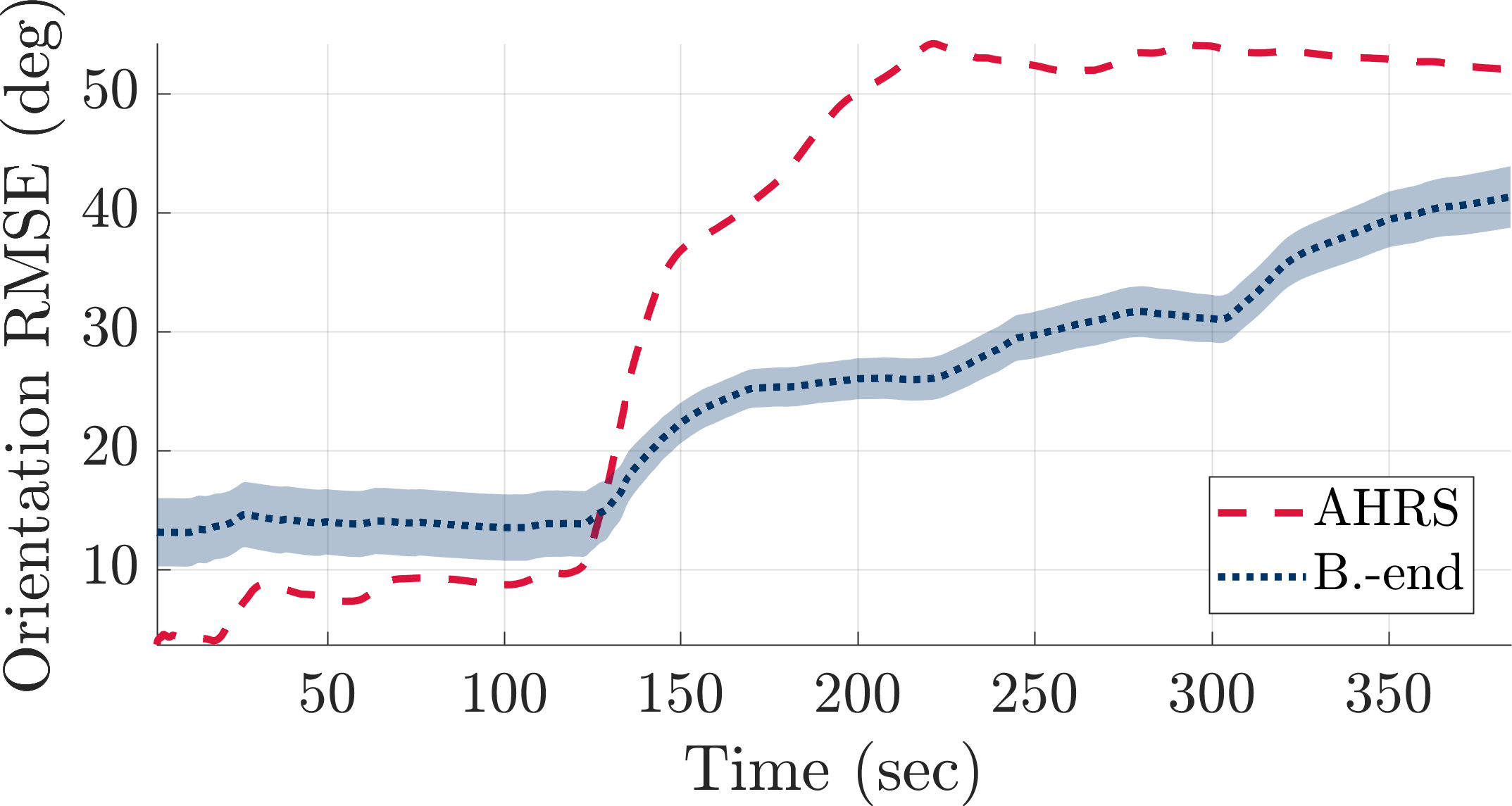}}~
\subfloat{\includegraphics[width = 0.66\columnwidth, trim={0.25cm 0.25cm 1.5cm 1cm},clip]{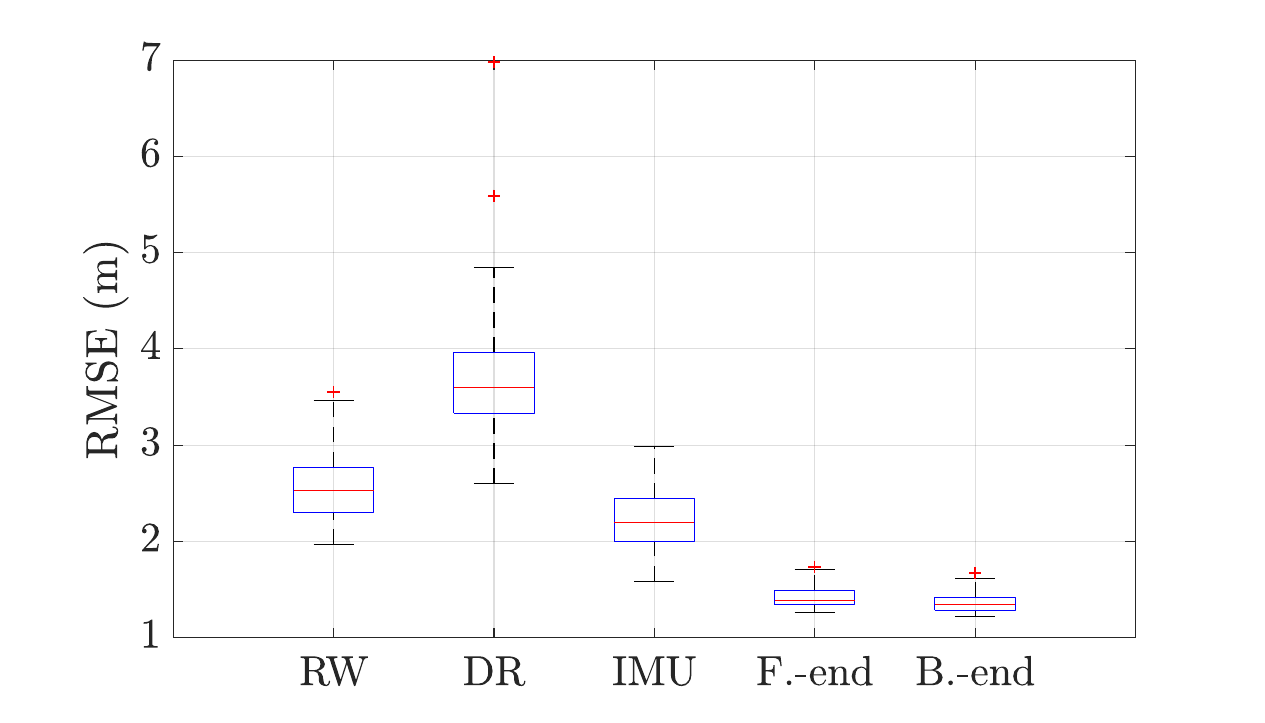}}
\caption{The evaluation results showing the effect of map perturbations on each technique. From left, respectively, the figures show position RMSE (with $99\%$ confidence bounds), orientation RMSE (with $99\%$ confidence bounds), boxplot showing the statistical summary of position RMSE from $100$ independent runs. The performance of the proposed system is near the case where an accurate prior map was given which confirms the system is robust to map errors.}
\label{fig:fx_rmse_maptest}
\squeezeup
\end{figure*}

\subsection{Localization Comparison under Accurate Map}
\label{subsec:mmcompare}
In this experiment, given an accurate map of the BLE beacons, we compare the performance of the localization techniques. We evaluate the results from $100$ independent runs on the collected dataset. Figure~\ref{fig:fx_results} shows the summary of results for all techniques. The evaluation consists of  Root Mean Square Error (RMSE) to show the performance of each technique as well as the empirical cumulative distribution function (CDF) of the position and orientation error. The position error is computed using the Euclidean norm of the three-dimensional position error. The orientation error is computed as the misalignment angle between the ground truth rotation matrix, $\mathbf{R}_{\mathrm{gt}}$, and the estimated rotation matrix, $\hat{\mathbf{R}}$, using $\lVert \log(\mathbf{R}_{\mathrm{gt}}^{\transpose} \hat{\mathbf{R}}) \rVert_{\mathrm{F}}$, where $\log(\mathbf{R})$ computes the matrix logarithm. The empirical CDF is an unbiased estimate of the population CDF and is a consistent estimator of the true CDF. Note that faster rise from zero to one along the vertical axis is a desirable outcome.

The proposed probabilistic IMU motion model improves the localization accuracy in all scenarios as it is a better proposal distribution to sample from. This model can predict the robot/device motion using IMU measurements, therefore, the drawn samples are more likely to be near the actual robot pose. Furthermore, the proposed radio-inertial localization and tracking system improves the overall system performance by decreasing the localization error.

However, the random walk motion model also provides comparable results and show that, given an accurate map, it can be a simple yet useful motion model. This model essentially explores the entire state space without any knowledge of actual actions. The DR motion model is based on counting motion increments; therefore, it cannot generalize the motion of the device. It is worth mentioning that by increasing the velocity noise variance in DR, the motion model's behavior approaches that of the random walk.

\subsection{Effect of Map Perturbation and Radio-inertial SLAM}
\label{subsec:rislam}
\begin{figure}[t!]
\centering
\subfloat{\includegraphics[width = 0.72\columnwidth]{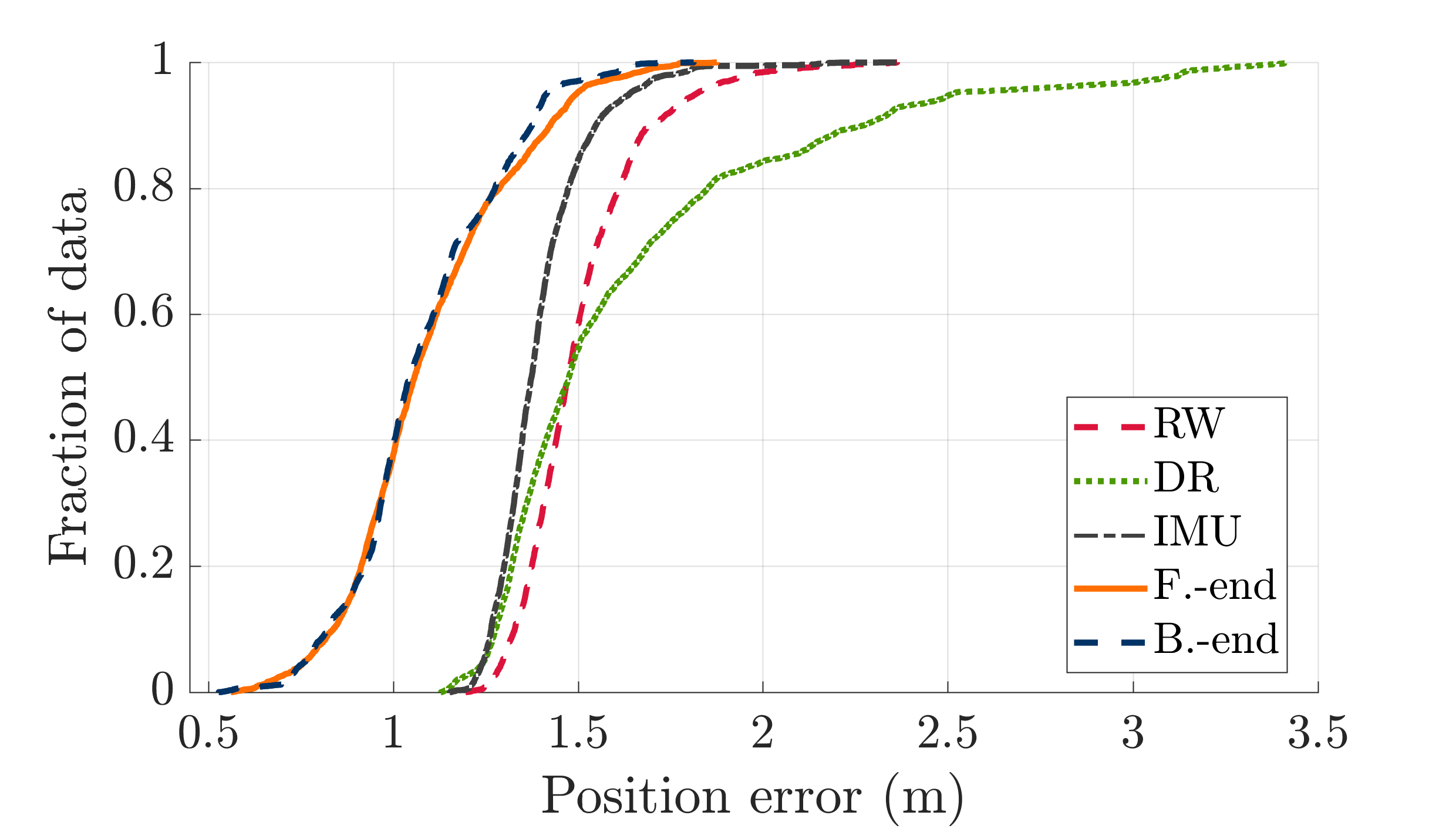}}\\ \squeezeup
\subfloat{\includegraphics[width = 0.72\columnwidth]{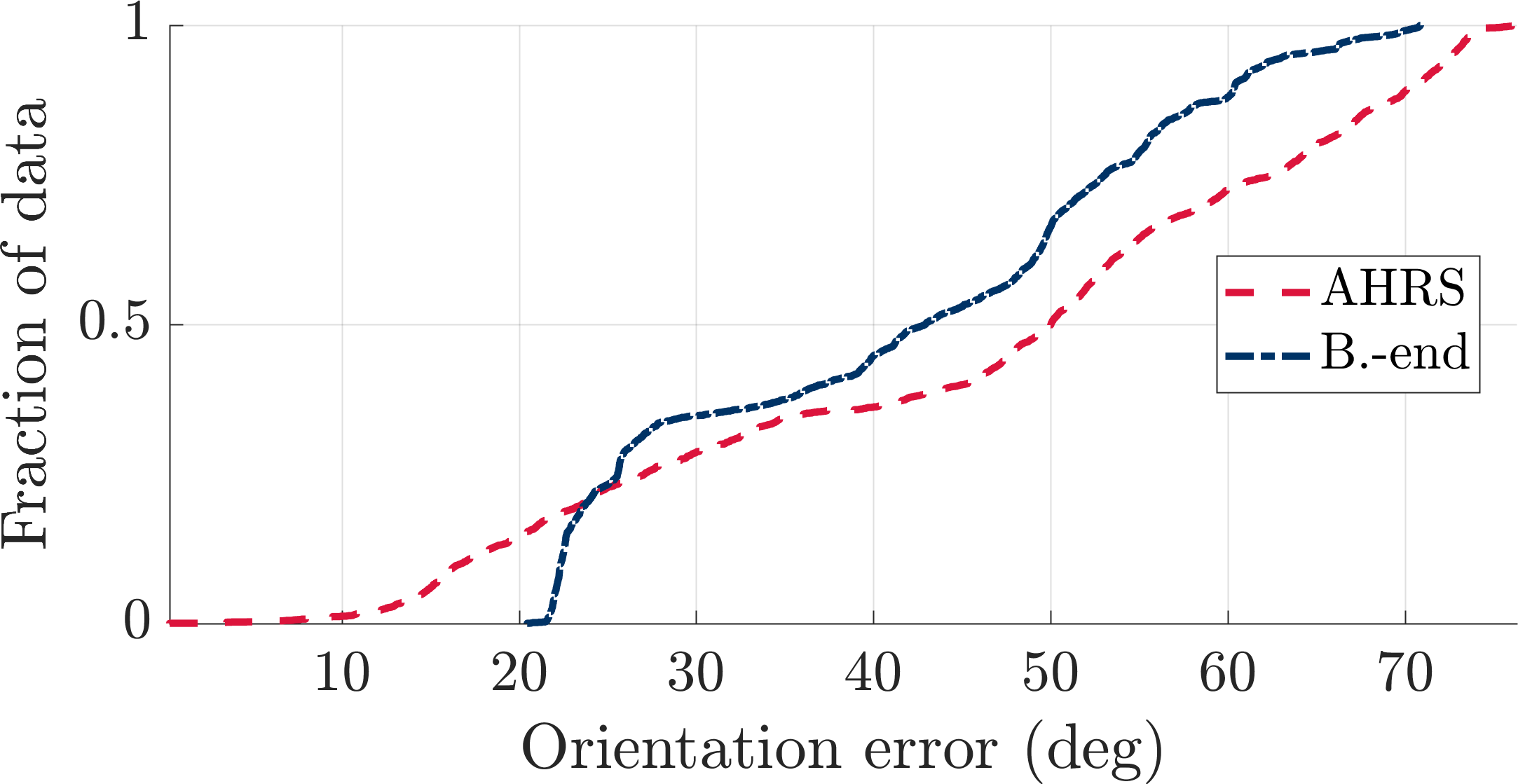}}
\caption{The empirical cumulative distribution functions of the compared algorithms under map perturbations. Each curve illustrates the median of $100$ CDF from $100$ independent runs.}
\label{fig:fx_cdf_maptest}
\squeezeup
\end{figure}

In the second experiment, we study the effect of large errors in the prior map of BLE beacons. We use a Monte Carlo simulation over $100$ independent runs using the same collected dataset. In each run, the location of each beacon is randomly perturbed by drawing noise from $\mathcal{N}(\mathbf{0}_3, 3 \mathbf{I}_3)$. This means the beacon can be located from near its initial position to $3\times 3 \m$ away from its true position along each axis. Figure~\ref{fig:fx_rmse_maptest} shows the position and orientation RMSE evolution over the experiment time as well as the statistical summary of position RMSE using a boxplot. Figure~\ref{fig:fx_cdf_maptest} shows the CDF plots of the position and orientation errors. The curves show the median of $100$ independent runs.

By definition, localization algorithms assume such a prior map is perfect; therefore, the expectation is to observe lower performance. However, in the closed-loop system,  where the Back-end solves the SLAM problem, the Back-end simultaneously solves for the beacons' map and the trajectory, while jointly estimating the system calibration parameters. The feedback to the Front-end results in better localization and tracking outcome from the PF (Front-end). Consequently, the more accurate position and velocity estimates from the Front-end provide better initial values for the Back-end optimization. Moreover, the proposed system does not discard any measurements and fuses all available data which reduces the estimation error.

Nevertheless, this test confirms that the proposed system, as expected, is robust to the error in the map. This property is highly desirable in practice since the exact measurements of the map can be challenging, or the map can be modified over time.

\subsection{Discussion and Limitations}
\label{subsec:limit}

\begin{figure}
\centering
\includegraphics[width = .8\columnwidth, trim={2.0cm 1.5cm 2.0cm 1.5cm},clip]{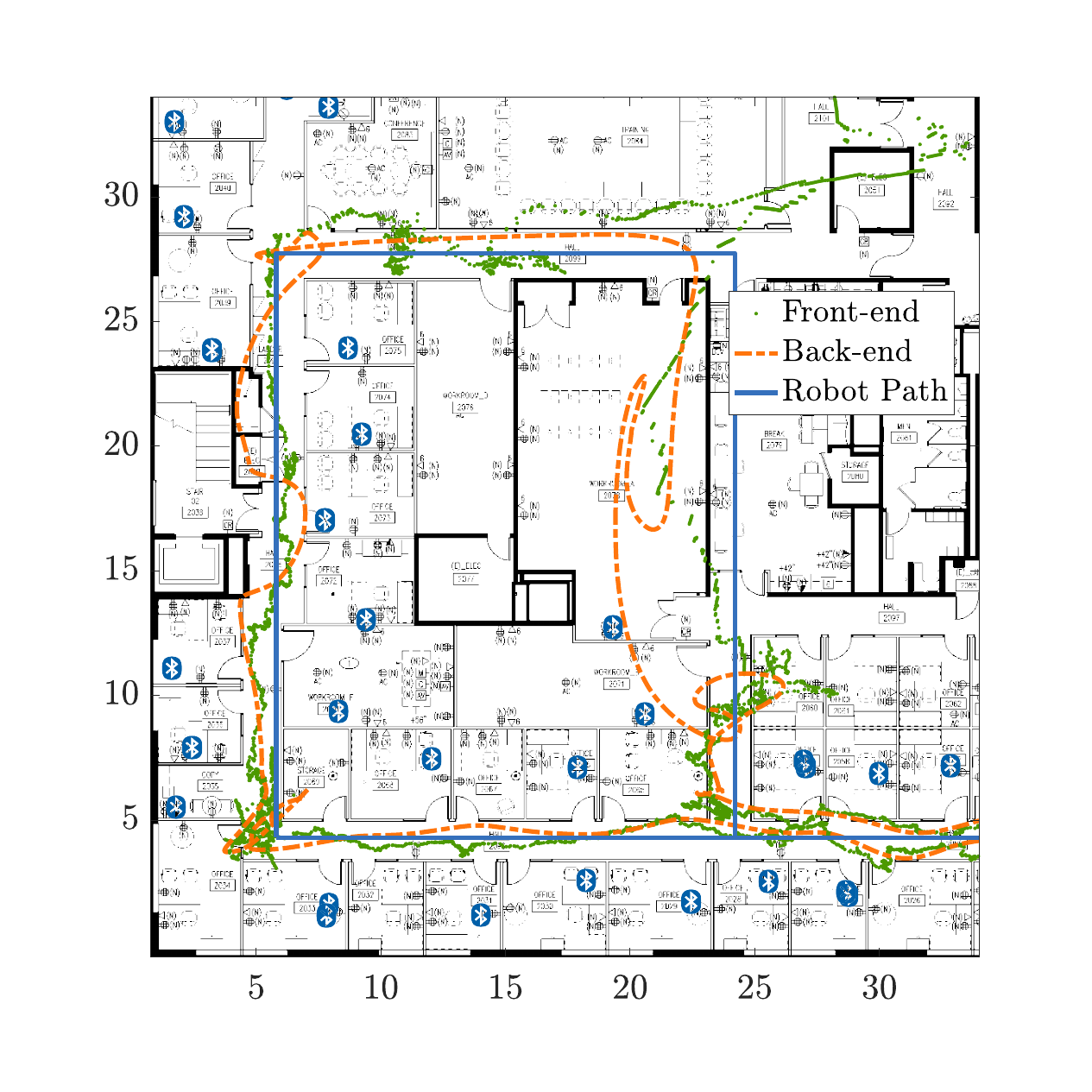}
\caption{The localization and tracking results in an office environment populated with BLE beacons. The Robot Path is only for the guidance.}
\label{fig:fx_office2}
\squeezeup
\end{figure}

The qualitative results on a larger dataset is shown in Figure~\ref{fig:fx_office2}. The robot starts from the bottom left corner and moves upwards. In this experiment, there are no beacons on the top and right sides of the rectangular path. Therefore, the system relies on its tracking ability. The error increases when beacons are sparse, as seen in Figure~\ref{fig:fx_office2}. The algorithm recovers the robot location as soon as the robot reaches near BLE beacons at about $(25,10)$ horizontal and vertical coordinates, respectively. Each beacon has a unique MAC address which solves the data association problem. Therefore, assuming there are sufficient beacons along the path, the proposed localization and tracking system error on average is fixed. A video showing the results is available on here:~{\footnotesize \url{https://www.youtube.com/watch?v=kEDGSnFvz8A}}

\section{Conclusion and Future Work}
\label{sec:conclusion}
We studied and developed a localization and tracking system that performs real-time and is robust to map errors. We developed a suitable motion model (proposal distribution) for the sequential Monte-Carlo algorithms that exploits the IMU dynamics to constrain the samples while improving the tracking ability. The proposed system has a closed-loop architecture and uses all available measurements for estimation.

Future work includes joint state and parameters estimation of radio signal measurements (RSSI factors) as done for IMU factors to extend the ranging to more than $10\m$. This approach enables the system to accept any type of RSSI signals, e.g.\@ WiFi or BLE, regardless of the transmission power. Adopting the idea in~\cite{jadidi2017gaussian} for robust observation selection can also be a step towards improving the system robustness as well as increasing the ranging to more than $10\m$. Finally, we think a \emph{visual-radio-inertial} SLAM technique is an attractive research direction to follow. The radio signals nicely complement visual measurements, while using visual data increases the estimation accuracy significantly.

\bibliographystyle{IEEEtran} 
\bibliography{refs}

\end{document}